\pdfoutput=1

\documentclass[11pt]{article}

\usepackage{acl}

\usepackage{times}
\usepackage{latexsym}
\usepackage{graphicx}
\usepackage{amsmath}
\usepackage{enumitem}
\usepackage{bbm}
\usepackage{tabularx}
\usepackage{tabulary}
\usepackage[T1]{fontenc}

\usepackage[utf8]{inputenc}

\usepackage{microtype}

\title{CUE Vectors: Modular Training of Language Models \\ Conditioned on Diverse Contextual Signals}

\author{Scott Novotney \and
    Sreeparna Mukherjee \and
    Zeeshan Ahmed \and
    Andreas Stolcke \\
  Amazon Alexa\\
  Seattle, WA, USA \\
  \texttt{\{snovotne,sreepar,ahzee,stolcke\}@amazon.com}}

\begin{document}
\maketitle
\begin{abstract}
We propose a framework to modularize the training of neural language models that use diverse forms of sentence-external context (including metadata) by eliminating the need to jointly train sentence-external and within-sentence encoders.
Our approach, contextual universal embeddings (CUE), trains LMs on one set of context, such as date and author, and adapts to novel metadata types, such as article title, or previous sentence.
The model consists of a pretrained neural sentence LM, a BERT-based context encoder,
and a masked transformer decoder that estimates LM probabilities using sentence-internal and sentence-external information.
When context or metadata are unavailable, our model learns to combine contextual and sentence-internal information using noisy oracle unigram embeddings as a proxy.
Real contextual information can be introduced later and used to adapt a small number of parameters that map contextual data into the decoder's embedding space.
We validate the CUE framework on a NYTimes text corpus with multiple metadata types,
for which the LM perplexity can be lowered from 36.6 to 27.4 by conditioning on context.
Bootstrapping a contextual LM with only a subset of the context/metadata during training retains 85\% of the achievable gain.
Training the model initially with proxy context retains 67\% of the perplexity gain after adapting to real context.
Furthermore, we can swap one type of pretrained sentence LM for another without retraining the context encoders, by only adapting the decoder model.
Overall, we obtain a modular framework that allows incremental, scalable training of context-enhanced LMs.
\end{abstract}

\section{Introduction}

Language models (LMs) estimate the prior probabilities of token sequences and are key probabilistic modeling components in a variety of applications, such as speech recognition, machine translation, or software keyboards.
When modeling linguistic token sequences, typical LMs model one sentence or utterance at a time, reflecting the fact that the strongest predictors of words are syntactic constraints and semantic associations within the sentence.
However, it has long been recognized that \emph{context} beyond the sentence has substantial influence on the word probabilities within a sentence.
Context literally means the surrounding text (or preceding text, when predicting words in temporal order), but can also refer to any extra-linguistic information, such as metadata (e.g., authorship, time, location) or associated other modalities (e.g., visual cues associated with a spoken utterance).

There is a large literature on leveraging such contextual information for language modeling, some of which we review below (Section~\ref{sec:prior_work}).
However, including context in language modeling presents major challenges for operational settings, especially when LMs need to be trained and deployed at scale:
\begin{itemize}
\setlength{\parskip}{0pt}
\item
    Context data is hard to come by. Many language corpora have no or very limited metadata, or contain unordered sentences that do not provide sequential context.
\item
	Context types are specific to a given source.  A newspaper corpus has metadata that is very different from spoken language data.
\item
	Use of context renders models context-specific, and therefore, less universally applicable.  With each type of context, a new model, or even model architecture, is required.
\item
	Context modeling requires more parameters, compute complexity and more training data.
\end{itemize}
All these difficulties lead to context being used sparingly in most practical settings, and only when it yields substantial benefits (such as in using a user's personal contact list in voice dialing).

In this paper, we propose a modular modeling framework for contextual language models, called contextual universal embeddings (CUE).
The fundamental idea is to separate the modeling of (1) sentence-internal LM, (2) context embedding and (3) combination of sentence-internal and contextual information each into their own modules.
First of all, we show that this architecture is an effective way to bring context to bear on the LM task, achieving 25\% relative perplexity reduction over a sentence-internal model, on a corpus of newspaper articles with rich metadata.
More importantly, each module can be trained separately, as opposed to jointly with the other modules.
Through experimentation we show that, for the practically important use-cases, training modules separately or incrementally preserves most of the achievable gain from contextual information.

Specifically, we can replace one type of context with another, while only adapting the context encoders to the new context, and retain 85\% of the best-case perplexity gain.
Maybe more surprisingly, we can train the decoder that combines context and sentence-internal information {\em without any actual context}, instead using noisy oracle unigram embeddings as a proxy. This recovers 67\% of the best-case gain after adapting to real context.  (Adapting the context encoders affects much fewer parameters, and takes much less data, than the model overall.)

Finally, we show that context encoders can be frozen and a whole different sentence-LM architecture swapped into the model ensemble.
After adapting only the combiner-decoder we obtain perplexity gains close to the optimum that would have been achieved by joint training of combiner and context embedding.

\section{Prior Work}
\label{sec:prior_work}
Longer text history is the most commonly used context in language models (LMs) \cite{mikolov2012context, jaech2018low, ji2015document, lin2015hierarchical}.
A naive way to bias a LM over text history is to ignore the sentence boundaries and train the contextual LM as the standard neural LM \cite{ji2015document}.
However, recurrent neural networks suffer training difficulties on longer sequences \cite{bengio1994learning} while transformer-style models are effective at incorporating this extra information  \cite{dai-etal-2019-transformer, NEURIPS2020_1457c0d6}.

Another approach is to summarize context into a single context embedding using a separate model.
For example, \newcite{mikolov2012context} and \newcite{pmlr-v32-le14} use topic information extracted from the context. 
\newcite{mikolov2012context} use a pretrained Latent Dirichlet Allocation (LDA) model while \newcite{pmlr-v32-le14} use paragraph embeddings learned during LM training.
\newcite{wang-cho-2016-larger} on the other hand, use a bag-of-words of whole text or individual sentences in the context to build the context vector.
\newcite{roh2020hierarchical} and \newcite{lin2015hierarchical} further extend sentence-based contextual models by using hierarchical embedding techniques.
This approach learns a representation of the context that is directly used as input to a neural LM.

Other sequence tasks in natural language processing (NLP) also leverage contextual information.
Neural machine translation (NMT) capitalizes on the availability of previous sentences on the source and target sides when translating documents \cite{Yun_Hwang_Jung_2020, sugiyama-yoshinaga-2021-context, zhang-etal-2018-improving}.
The only difference in the approaches is how the context is encoded into a representation optimized for NMT.

Automatic speech recognition (ASR) and conversational dialog systems also use contextual information, such as recent advances in shallow or deep fusion of end-to-end neural architectures  \cite{zhao19d_interspeech, 47384, kim2018dialog,munkhdalai2021fast,jain2020contextual}.
Recent papers have also considered biasing LMs with context beyond the previous sentence and incorporate additional signals such as date-time, geolocation or gender \cite{8461979, diehl-martinez-etal-2021-attention} or application metadata like dialog act or intent \cite{masumura19_interspeech, shenoy2021contextual, 7953251}.
Other sources of context used to bias LMs are personalized content \cite{jaech-ostendorf-2018-personalized, fiorini-lu-2018-personalized}; conversational turn-taking \cite{xiong-etal-2018-session}; multi-modal sources \cite{8639551}; or even user demographics to suggest fashion suggestions \cite{denk-peleteiro-ramallo-2020-contextual}.
 
\section{Architecture}
Our task is to estimate an auto-regressive language model conditioned not only on the previous words in the sentence, $W=w_1,w_2,\ldots,w_n$, but also on several contextual signals,
\begin{equation}
  P(W | C) = \prod_{i=1}^n{P(w_i|w_{i-1}, w_{i-2},\ldots, C)}
  \end{equation}
where $C=[c_1, ..., c_K]$ represents a set of $K$ contextual inputs that may vary with each sentence, and where each $c_k$ is a sequence of tokens from the same vocabulary as the target sentence.

\begin{figure}[tb]
\centering
\includegraphics[width=\columnwidth]{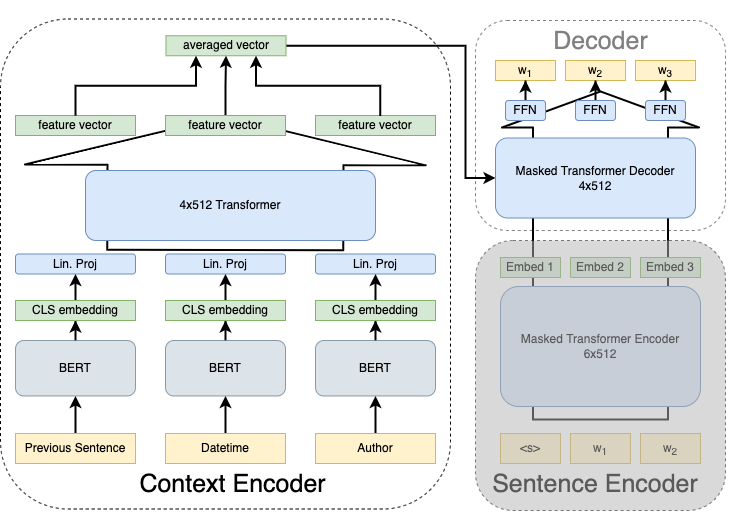}
\caption{ {\em Overview of CUE architecture.}  Pretrained sentence encoder and DistilBERT modules are frozen.}
\label{fig:arch}

\end{figure}

Adapting recent work in hierarchical contextual embeddings \cite{Yun_Hwang_Jung_2020}, our CUE architecture has three components (see Figure \ref{fig:arch}):
\begin{itemize}
\setlength{\parskip}{0pt}
\item An auto-regressive transformer \textbf{sentence encoder} conditioned on within-sentence history.
\item A transformer \textbf{context encoder} with a BERT-based embedding combines multiple context signals into one CUE vector.
\item An auto-regressive transformer \textbf{decoder} to predict the current word based on the context and within-sentence embeddings.
\end{itemize}

Our goal is to train a separate context encoder that can be updated without modifying the sentence encoder or decoder and run inference independently of the other modules.
This is important for operational practicalities, to precompute context embeddings and update them incrementally without requiring downstream components to change.

\subsection{Context Encoder}
We present all the contextual signals to the context encoder as strings.
Non-textual signals, like datetime, are converted into English, such as “Wednesday 29 May 1985”.
Similarly, we represent any categorical or symbolic context by its English text string.

The encoder projects the set $C$ of $K$ context types into one compact embedding.
These may consist of the previous $K$ sentences, $K$ different metadata types such as datetime, or a mix of both.
We then encode each context string $c_k$ with DistilBERT \cite{Sanh2019DistilBERTAD} and represent each context by its \texttt{CLS} embedding to generate the intermediate representation
\begin{equation}
g_k = FFN(BERTEncoder(c_k)).
\end{equation}
We do not fine-tune DistilBERT since empirically it gave negligible gains on our experimental corpora.

The set of intermediate representations $G=[g_1; \ldots; g_K]$ is then passed through transformer blocks to learn dependencies between the context types and to generate the self-attended embeddings $E=[e_1;\ldots;e_K]$, where 
\begin{equation}
E = TransformerEncoder(G).
\end{equation}
The contextual encoder is invariant to the ordering of context types since we treat context as a ``bag of sequences'' and do not add positional embedding to the \texttt{CLS} embeddings.
Additionally, we do not use query values from the within-sentence encoder, so as to preserve the modularity of our architecture; our goal is to use one context encoder with multiple sentence encoders or modeling tasks.
The empirical gain was small for conditioning the context attention on the history at each word position (thus giving a different context vector for each token).

Finally, the per-context embeddings $e_k$ are averaged to produce our compact representation,
\begin{equation}
e_{\text{cue}} = \frac{1}{K}\sum_{k} e_k .
\end{equation}

\subsection{Sentence Encoder}
The sentence encoder is a familiar auto-regressive masked language model with a transformer encoder and a final softmax layer to generate a distribution over the vocabulary \cite{attention_all_you_need}.
We used six layers of 512 dimensions each with 4 attention heads and used a standard language modeling task to fit the parameters; no context was used to train this module.
In our experiments, the sentence-encoder parameters are frozen and never fine-tuned when biasing the decoder with context. 
We assume that the sentence encoder was trained on a very large general text corpus.
It uses the same DistilBERT tokenizer as the context encoder, but do not use DistilBERT for word embeddings since our model is causally auto-regressive.

\subsection{Decoder}
The decoder is a masked transformer decoder as described in \newcite{attention_all_you_need} with six layers of 512 dimensions and 4 heads for multi-headed attention.
The sentence-internal embeddings (before the softmax layer) are passed as the shifted outputs to the decoder; 
along with the contextual CUE vector $e_{\text{cue}}$ as input to the multi-headed attention module in the decoder.

\section{Adapting to Evolving Context}

\begin{table}[tb]

\centering
\small
\begin{tabular}{l|r|c|c}
{\bf Module } & {\bf \# Params} & {\bf Train } & {\bf Adapt. } \\
\hline
Sentence encoder & 24.5M & N & N \\
DistilBERT & 65M & N & N \\
Decoder & 40M & Y & N \\
Context encoder & 6.6M & Y & Y \\
Total & 	136M & - & - \\

\end{tabular}
\caption{{\em CUE components.} Some modules are updated or frozen depending on context training or adaptation.}
\label{model_params}
\end{table}

We now no longer assume that the set of context types is static between training and test.
For example, an API providing context may be retired; or business rules improving customer privacy may remove geographic information.
The set of contexts may evolve over the life-cycle of our CUE encoder and we now introduce an adaptation step.%
\footnote{Out of scope for this paper is missing context at inference time. We assume the same set of contexts at adaptation and testing time.}

Ideally, we would jointly fine-tune the entire model architecture (context encoder, sentence encoder and decoder) on annotated data that contains the new context types.
However, this creates an operational burden since different downstream decoders that use context embeddings would each need retraining.
Our goal is to adapt the CUE context encoder while leaving the decoder parameters frozen.
This will minimize the number of parameters to be retrained and simplify model deployment by factoring the context encoder from the decoder.

We break the training process into two phases:
{\em Training} constructs the initial set of model parameters and is not constrained by operational needs.
{\em Adaptation} happens at some later point in time after the set of context types changes.
Section \ref{sec:adapt1} considers the scenario where new context types are added to or replace the initial training types.
Section \ref{sec:adapt2} assumes {\em no} context is available during model training, only at adaptation time.

\subsection{Adapting with annotated data}
\label{sec:adapt1}
Our adaptation strategy is to fine-tune only the context encoder, leaving the other parameters unchanged.
Since the context encoder consumes sequences of text, our approach benefits from DistilBERT projecting sentences into a shared embedding space through the \texttt{CLS} token prepended to the beginning of the sentence.
We fine-tune the context transformer that operates on the per-context DistilBERT embeddings before averaging (see Figure \ref{fig:arch}).
This component has 6.6M parameters, roughly 5\% of the total number of parameters (see Table \ref{model_params}).
Given an adaptation corpus of sentences paired with new context types, we take a forward pass for each batch and then backpropagate through the decoder to the contextual encoder.
This approach handles any arrangement of new context.
Section \ref{sec:adaptation} details results evaluating how adaptation benefits over zero-shot approaches with a static model.

\subsection{Proxy embeddings}
\label{sec:adapt2}
We now consider the scenario where we have no sentences paired with context during training, but want to bias our architecture on context at adaptation time.
Since the decoder parameters are frozen during adaptation, we must prime it during training to pay attention to a context embedding, even though we lack the context to generate such an embedding.
We tackle this problem by hypothesizing that context plays a role similar to a topic model: it mostly affects the unigram distribution, with small higher-order effects.
Thus, we generate proxy embeddings from an oracle encoding the unigrams in the target sentences, described below.

\begin{figure}[tb]
\includegraphics[width=\columnwidth]{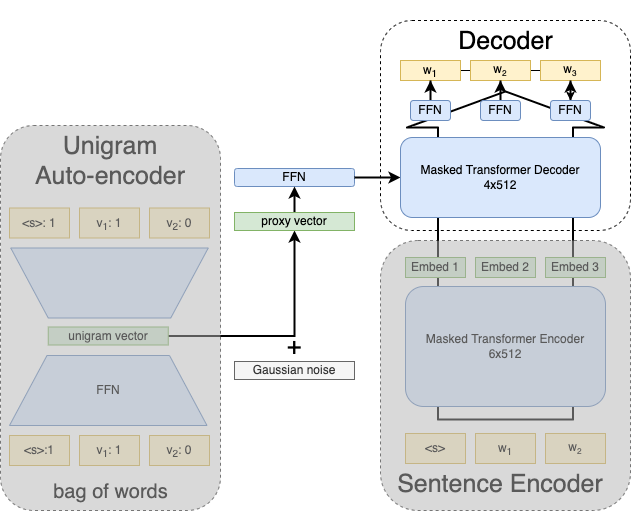}
\caption{{\em Priming the decoder with proxy embeddings}. We add noise to an embedding of the target sentence unigram distribution as a proxy for the decoder to learn to attend to context as yet unknown during training. Modules in gray are frozen during decoder training.}
\label{fig:uni}
\end{figure}

\newcommand{\stimes}{{\times}}
\newcommand{\splus}{{+}}

\subsubsection{Generate unigram embeddings}	
We first transform each sentence $W=w_1,\ldots,w_n$ in our training corpus $\mathcal{D}$ into an empirical unigram distribution (``bag of words'') over the vocab $\mathcal{V}$, 
\begin{equation}
\tilde{P}(W) = \left\{ w \in \mathcal{V} :  \frac{\sum_{i=1}^n \mathbbm{1}(w_i = w)}{n} \right\}.
\end{equation}
Next, a feed-forward auto-encoder, $\mathcal{F}_\Theta$, reconstructs $\tilde{P}(W)$ through a low-dimensional hidden layer 
fitted by minimizing reconstruction loss with Kullback–Leibler (KL) divergence,
\begin{equation}
\mathcal{L}(\Theta ; \mathcal{D}) = \sum_{i=1}^N \text{KL}(\mathcal{F}_\Theta(\tilde{P}_i)|| \tilde{P}_i).
\end{equation}
The layers were $28996 \stimes128\stimes16\stimes128\stimes28996$ with ReLU non-linearities, and a final softmax layer to generate a distribution over the vocabulary.
We swept multiple architecture sizes and saw no gain for more parameters.
Reconstruction loss on the test set improved from 5.59 to 1.94 after ten epochs.

\newcommand\norm[1]{\left\lVert#1\right\rVert}
\newcommand\proxy{\hat{a}}

\subsubsection{Train decoder with proxy embeddings}
We then replace the context encoder with this auto-encoder; freeze the sentence encoder and fit the parameters of the decoder on training data that do not have context annotations (see Figure \ref{fig:uni}).
In place of the context embedding, we construct a proxy embedding $\proxy$ by adding Gaussian noise to the embedding of the {\em entire} sentence and re-normalizing.
\begin{align}
\proxy & = \mathcal{F}_\Theta (\tilde{P} ) + \mathcal{N}(0,\sigma^2) \\
\proxy &= \frac{\proxy}{  \norm{ \proxy }^2},
\end{align}
As we increase $\sigma$, the information content in $\proxy$ decreases, calibrating the information content of the proxy embeddings to match the expected strength of the actual context.
Section \ref{sec:adaptation} details the importance of this hyperparameter.
We then project this low-dimensional embedding to the target contextual embedding (512 in our experiments) through a linear projection and pass it as input to the decoder.

\subsubsection{Adapting the context encoder}
Once annotated sentences with context are available for adaptation, we train only the context encoder to project available context into an embedding space tuned to the decoder.
The decoder was ``primed'' to attend to an external embedding and the sentence-internal embeddings.
We freeze the decoder and sentence encoder weights; backpropagate; and update only the weights of the transformer in the context encoder and the linear projections that scale from DistilBERT embeddings to the context embedding dimension.
As mentioned above, our encoder is agnostic to the ordering of the context types and transforms text into an embedding through DistilBERT.
Section \ref{sec:experiments} demonstrates that this approach successfully adapts to unseen context data.

\section{Corpus}
\label{sec:corpus}
We used the New York Times Annotated Corpus \cite{sandhaus2008new} released through Linguistic Data Consortium (catalog number LDC2008T19) containing over 1.8M English articles spanning 1987 to 2007.
This corpus includes a rich collection of contextual annotations for each article, ideal for evaluating our CUE framework.
Each article contains up to 47 different metadata types that were labeled by humans (author, title, desk) or algorithmically  (locations, topic).
We down-selected from 47 to 11 distinct metadata signals after removing redundant or uninformative context.
All context types were character sequences and include previous sentence, title, author, entities present in the article, section descriptors, date, and topic descriptors  (see Appendix \ref{app:context_stats} for details).
Articles averaged 32 sentences in length and average sentence length (after tokenization) was 26.
We trained the sentence encoder on a large subset of articles; used separate training and adaptation corpora and separate validation and test sets (Table \ref{table:stats}).

\begin{table}[tb]
\resizebox{\columnwidth}{!}{
\centering
\begin{tabular}{l|r|r|r}
\bf Purpose & \bf \#Articles & \bf \#Sentence & \bf \#Words \\
\hline
Word LM training  & 250K & 8.5M & 215M\\
Context training & 55K & 1.8M & 47M\\
Context adaptation & 60K & 2M & 41M\\
Validation & 5K & 170K & 4.3M\\
Test & 5K & 166K & 4.2M\\
\end{tabular}
}
\caption{{\em NYTimes corpora used in this work.} We randomly shuffle all articles before partitioning and use 20\% of the entire corpus to reduce experiment turnaround time.}

\label{table:stats}
\end{table}

\section{Experimental Results}
\label{sec:experiments}
\subsection{Hyper-parameters}
Sentences were tokenized first with spaCy \cite{spacy} and then into word pieces using the DistilBERT tokenizer.
We evaluated model performance by computing perplexity (PPL) on the heldout test set.
We trained all models for ten epochs using the AdamW \cite{adamw} optimizer and One Cycle learning rate scheduler \cite{onecycle} with a learning rate of 0.0001, maxing at 0.004; and gradient clipping of 0.95.
We parallelized batches on 8 V100 GPUs and averaged 75k tokens per second with a per-GPU batch size varying between 64 and 256 sentences depending on the architecture size.
The parameters of the sentence encoder and DistilBERT are frozen for all the experiments, greatly speeding up training time with negligible impact on PPL.

\subsection{Contextual biasing}

\label{sec:baseline_exp}

\begin{table}[tb]
\centering
\small

\begin{tabular}{l|c|r}
{\bf Contextual features} & {\bf Test PPL }& {\bf Rel.\ PPL} \\
\hline
Sentence-internal only & 36.6 & - \\
+ article metadata & 35.9 & -2.0\% \\ 
+ previous sentence & 29.8 & -18.6\% \\
+ previous sentence + metadata & 27.4 & -25.0\% \\
\end{tabular}
\caption{ {\em Reduction in PPL by adding context.}
We contrast a sentence-internal transformer LM with four variations of added contextual information.
Article metadata (e.g., author, title) is mildly informative, the previous sentence is the most useful. Metadata improves PPL {\em more} when previous sentence is included.
}
\label{table:baseline}
\end{table}

We first compare our architecture against a sentence-internal auto-regressive language model.
The 6x512, 4-head transformer word language model was trained on the separate 200M-word corpus and used as the sentence-encoder in our full CUE framework.
As shown in Table \ref{table:baseline}, contextual signals reduce PPL by 25\% for this corpus and nearly three fourths of that gain is due to the previous sentence.
Since the remaining contextual features are at the {\em article} level, they have a smaller impact on within-sentence likelihoods.%
\footnote{See Appendix \ref{app:feats} for a breakdown of the relative strength of each contextual type.}
This 25\% relative gain is the upper bound for adaptation methods since context types are consistent between training and test; and context encoder and decoder are trained jointly.

To evaluate the key elements of our architecture, we conducted an ablation study by disabling various components and measured the relative degradation in perplexity, as shown in Table \ref{table:ablation}.
Removing the transformer after DistilBERT embeddings and using a simple average gives an 8\% degradation.
Removing the transformer decoder and instead concatenating the CUE vector with each step's hidden state before the logit layer gives a 22\% degradation.
Replacing DistilBERT with a randomly initialized transformer estimated on the contextual training corpus give the biggest loss of 27\%.
Finally, using only the context to predict each word (a constant vector at each step) is much worse, but still 45x better than random (which would be equal to the vocab size of 28,996).
Context is a useful prior, even though it is constant for all tokens in the sentence.

\begin{table}[tb]
\centering
\small
\begin{tabular}{l|c|r}
{\bf Module } &{\bf Test PPL} & {\bf Rel. PPL } \\
\hline
Full architecture & 27.4 & 0\% \\
No context transformer & 29.6 & +8\% \\
No decoder transformer & 33.5 & +22\% \\
No DistilBERT & 34.8 & +27\% \\
No contextual inputs & 36.6 & +33\% \\
No sentence inputs & 643 & +2200\% \\
\end{tabular}
\caption{{\em Ablation study on architecture modules.} Refer to Figure \ref{fig:arch} for a schematic of the components.
}
\label{table:ablation}
\end{table}

\begin{figure}[tb]
\includegraphics[width=\columnwidth]{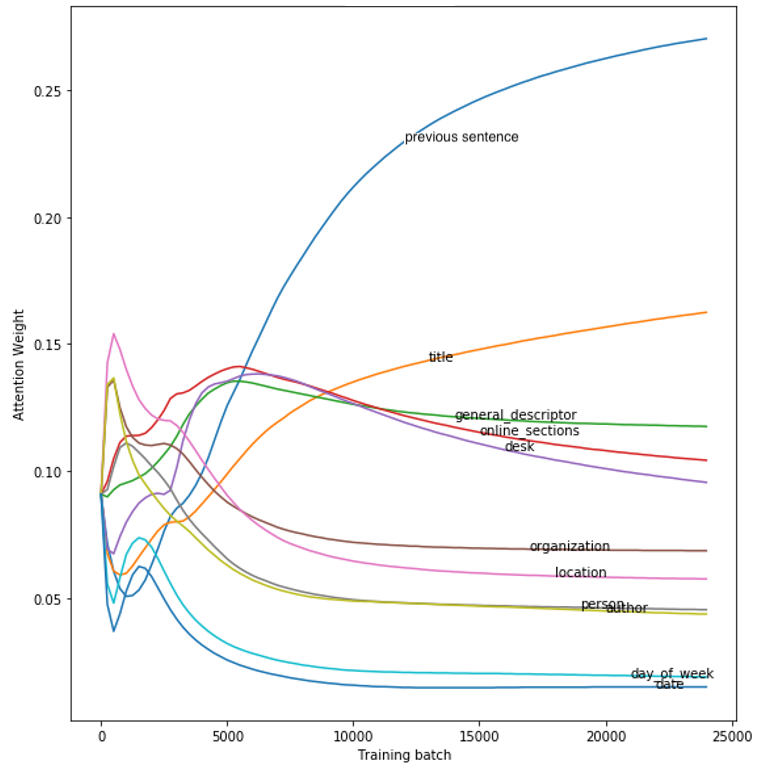}
\vskip -0.3cm
\caption{{\em Change in normalized attention weights over training iterations. }
The weights of the self-attention component of the context encoder converge to the relative importance of each contextual category over time, with previous sentence receiving the most weight. 
}
\label{fig:attn}
\end{figure}

Figure \ref{fig:attn} captures the model's attention converging to the relative importance of each context type.
The ordering of context types by attention weights is similar to a ranking by perplexity gain given in Appendix \ref{app:feats}.

\subsection{Adaptation}
\label{sec:adaptation}

\newcommand\A{{\bf A}}
\newcommand\B{{\bf B}}

We next evaluate our framework for interchangeability of different forms of context.
We randomly partitioned the eleven context types into two sets \A\ and \B\ and report the average over five separate trials in Table \ref{table:adapt}.
Set \A\ is our training set and we experiment with two adaptation scenarios: \B\ replaces \A\  or \B\ is added to \A.

\begin{table*}[t]\centering
\centering
\small
\begin{tabular}{c|l|c|c|c|l|l}

\bf Row &  {\bf Description} & {\bf Train } & {\bf Adapt } &   \bf Test & {\bf PPL} & {\bf Rel. PPL } \\
\hline

1& Word-only baseline         & - & -  & - &36.6 & -  \\
2& Proxy training (cheating) & Proxy & -  & Proxy & 29.7 & $-19\%$ \\
3& Context A always available           & A        & -     & A & $29.3 \pm 1.3$  &  $-20 \pm 4 $\% \\
\hline
4& No training context        & Proxy  & B   & B & $31.8 \pm 0.8$  &  $-13 \pm 2$\% \\
5& No adaptation                & A           &  -      &  B & $32.0 \pm 0.2 $ & $ -13\pm 1$\% \\
6& Context A replaced by B              & A        & B   &  B& $30.9 \pm 0.7 $ &  $-16 \pm 2 $\% \\
7& Context B always available          & B         & -  &  B & $29.9 \pm 1.4$ &  $-19 \pm 4 $\% \\
\hline
8& No training context        & Proxy     & A+B  & A+B &30.4  & $-20$\% \\
9& No adaptation                & A           &  -      & A+B &  $30.1 \pm 2.1 $ & $-18 \pm 6$\% \\
10& Context B added after training    & A           & A+B   & A+B &$28.8 \pm 1.0 $  &  $-21\pm 3$\% \\
11& Context A+B always available     & A+B       & - & A+B   &  27.4 &$ -25$\% \\
\end{tabular}
\caption{ {\em Adaptation results.} Results are averaged over five random partitions of context types into training set \A\ and adaptation set \B. Results without std.~dev.\ are based on a single experiment run. Adding metadata with CUE embeddings outperforms a word-only model (row 1) by 25\% (row 11). CUE vectors are robust to evolving context, either without any context in training (rows 4, 8);  no adaptation (rows 5, 9); or adapting with new annotated sentences (rows 6, 10). Contrast with lower bound of all context available in training and adaptation (rows 7,11). }
\label{table:adapt}
\end{table*}

When {\em adding} additional context types ($\A\rightarrow \A\splus\B$), adapting the context encoder without retraining the decoder captures 85\% of the possible gain for jointly training the context encoder and decoder on all context types (compare rows 1, 10 and 11).
Starting at a word-only PPL of 36.6, adaptation to $\A\splus\B$ reaches 28.8 versus the lower bound of 27.4.
When {\em replacing} context types ($\A\rightarrow \B$), adaptation also achieves 85\% of the possible gain (36.6 to 30.9 versus the lower bound of 29.9 in rows 1, 6 and 7).
 Even without {\em any} adaptation (rows 5 and 9), our architecture generalizes to new context types (approximately 70\% of the possible gain), though not as effectively as with adaptation.
This is because we transform context into English text and leverage BERT embeddings as a strong initial embedding for context sequences.

\begin{figure}[tb]
\includegraphics[width=\columnwidth]{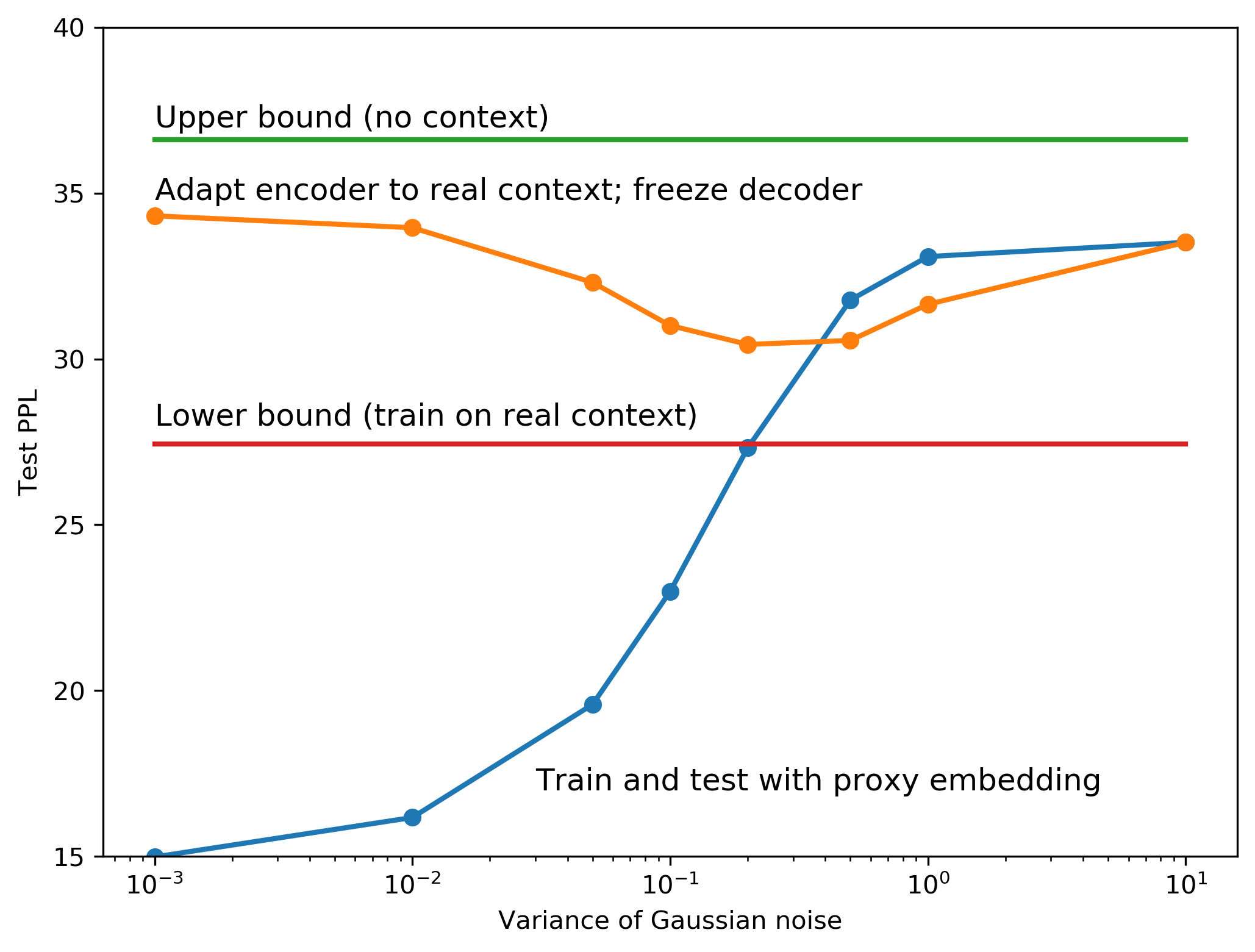}
\vskip -0.3cm
\caption{{\em Varying proxy embedding strength.}
The baseline is no context (green line) versus the lower bound of knowing all context in training and test (red line).
We sweep the amount of noise added to the oracle unigram vector on the x-axis.
When training and testing on only the unigram vector (blue line) the unigram vector is a powerful oracle without noise, but then becomes random as the variance increases.
During adaptation (orange line), we discard the unigram embeddings, freeze the decoder parameters, and retrain the context encoder (5\% of parameters).
The amount of embedding noise is optimal roughly when the proxy embedding is as informative as actual context (where blue and red lines intersect).}
\label{fig:proxy_sweep}
\end{figure}

When we train the decoder with proxy embeddings (no real context at all) and adapt to context, the PPL is within 6\% to 11\%  (depending on the context subset) of the lower bound of jointly training the context encoder and decoder.
This approach recovers  67\% of the gain from jointly training context encoder and decoder for the two scenarios ($\A\rightarrow\B$ and $\A\rightarrow\A\splus\B$).
We find this quite remarkable given that the decoder knows nothing of context during training; the result validates our hypothesis that context encodes the unigram priors.

We tuned the strength of the proxy embedding by sweeping the variance of Gaussian noise added.
The sweet spot is where the information content in the proxy embedding is close to the actual context, as shown in Figure \ref{fig:proxy_sweep}.
This intuitive result provides a sensible recipe for setting this hyperparameter in a production setting.

\subsection{Different sentence encoders}
Our CUE architecture factors the context encoder from the sentence encoder and decoder.
This approach generates one embedding that can be used  with multiple decoders and sentence encoder pairs.

To evaluate the generalizability of our CUE framework, we trained a 4x512 LSTM sentence encoder and froze its model parameters for the remaining experiments.
We then trained a new decoder using the LSTM sentence encoder and evaluated two different context encoders:
1) randomly initialized and jointly trained with the decoder {\em or}
2) the pretrained encoder jointly trained with the old, {\em transformer}-based sentence encoder and frozen parameters.
Table \ref{table:lstm} shows that the CUE vectors trained with one sentence-LM architecture are useful to the other, with a relative degradation when swapping of 7\% and 1\%, respectively, between LSTM and transformer sentence encoders.

\begin{table}[htb]
\center
\small
\begin{tabular}{l|l|l|r}
\bf Row & \bf Sentence Encoder & \bf Context Encoder & \bf PPL  \\
\hline

1 & Transformer, frozen & jointly trained  & 27.4 \\
2 & Transformer, frozen & frozen from (3) & 29.4 \\
\hline
3 & LSTM, frozen & jointly trained & 28.0 \\
4 & LSTM, frozen & frozen from (1) & 28.2 \\
\end{tabular}
\caption{{\em Swapping context and sentence encoders.} Without fine tuning, frozen context encoders generalize to new sentence encoders and decoders with <7\% relative degradation (compare  rows 1 and 2; 3 and 4).}
\label{table:lstm}
\end{table}

These results suggest that our CUE framework can factor the context encoder and decoder training and generalize to multiple decoder architectures.
This frequently occurs in operational settings such as the first and second pass LMs in speech recognition or compressed parameter sizes due to memory and latency constraints.

\section{Discussion}
We analyzed whether a sentence's context behaves like a cache model, since it contains textual data from the previous sentence, title, and other contexts.
To better understand this effect, we divided test data tokens into two bins: Those that appeared in the text of the sentence's context and those that did not.
We measured the relative gain in log likelihood when conditioning the LM on context versus not.
30\% of the tokens appeared in the context (cache) and the relative gain was 74\%---there clearly is a strong benefit for recurring tokens and the CUE encoders capture this effect.
The 70\% of tokens that do not occur in the cache improved their log likelihood by 26\% relative.
So the cache effect does not explain the entire benefit of CUE vectors and correlations among different token types are captured as well.
The top context types that had tokens in the sentence were previous sentence (23\%), title (9\%) and person (6\%).

To verify that the empirical improvements from the previous sections are semantically plausible, we analyzed the context embeddings of the {\em first sentence } of 5000 heldout articles.
These embeddings do not contain information from the previous sentence and thus represent the entire article's metadata.
Figure \ref{fig:tsne} projects these embeddings down to two dimensions with t-SNE.
We then clustered the vectors with k-means and aggregated word counts for all articles within a cluster.

Finally, we display the five most salient words (computed with TF-IDF) from the context and, separately, from the article text.
Even though the articles were clustered based only on context, the groupings of {\em article text} are semantically meaningful, with clear clusters related to newspaper sections such as corrections, marriage announcements, sports and other news related topics.
Our context embedding is preserving semantic information.

One limitation of the proxy embedding approach is that they may not extend to other NLP tasks, like named entity tagging.
Since they are derived from the unigram embedding, they directly encode the targets of the language model task.
This may not prove useful for higher-order annotations and further work should look into a multi-task proxy embedding that directly optimizes an ``interface'' embedding space instead of a unigram distribution.

\section{Conclusions}
\label{sec:conclusion}
We introduced the CUE framework to factor context encoding and next word prediction of context-aware neural language models.
Unlike previous work, we do not assume that the set of context signals is constant between training and test.
We optimize the model architecture to reduce the operational burden of managing and retraining of large neural LMs over their life cycle.

Our approach is robust to changing context types; by adapting only 5\% of the parameters, we recover 85\% of the possible gain from jointly training all components.
Furthermore, we introduce {\em proxy embeddings} to pretrain a decoder to be attuned to external context embeddings even when those are not known at training time.
This approach is 67\% as good as jointly training with all context.

In future work, we would like to handle missing context at inference time through data imputation or dropout approaches.
Furthermore, we plan to extend the proxy embedding approach such that the context encoders can be trained fully independent of the decoder.

\vfill
\pagebreak

\bibliography{prior_work}
\bibliographystyle{acl_natbib}

\vfill
\clearpage
\appendix

\onecolumn

\section{Appendix: Examples of Context Types}
\label{app:context_stats}

\begin{table}[h]
\settowidth\tymin{\textbf{Avg. token length}}
\setlength\extrarowheight{2pt}
\begin{tabulary}{1.0\textwidth}{L|L|R|R}
\textbf{Type} & \textbf{Example} & \textbf{Avg.\ \#\ tokens} & \textbf{\%Articles}\\
\hline
Author & Michael T. Kaufman & 6 & 84.0\%\\
Date & May 25 1985 & 4.6 & 99.9\%\\
Day of Week & Monday & 1 & 99.9\% \\ 
Descriptor & Computers And The Internet & 5.1 & 83.4\% \\
Desk  & Business/Financial Desk & 12 & 99.5\%\\
General Descriptor & Surfing, Ranching & 4.5 & 100\% \\ 
Location & New York, NY & 4.5 & 42.1\%\\
Online Section & Business; Technology & 4.5 & 99.0\%\\
Organization & Linguistic Data Consortium (LDC) & 5 & 42.4\%\\
Person & Bloomberg, Michael & 8 & 83.7\%\\
Previous Sentence & beloved wife of the late freddy pomerantz.   & 26 &97.0\%\\
Title & Voice Recognition Is Improving, but Don't Stop the Elocution Lessons & 8& 100\% \\
 \hline
\end{tabulary}
\end{table}

\section{Appendix: Relative strength of context types}
\label{app:feats}

\begin{figure}[h]
\centering
\includegraphics[width=0.6\columnwidth]{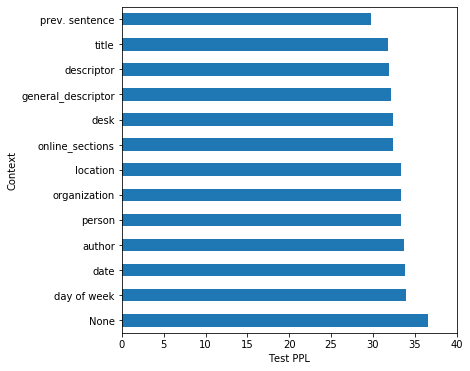} 

\caption{{\em Relative strength of each contextual type}. We trained the CUE model with only one contextual signal at a time and measured perplexity on the same heldout test set. Textual context types (previous sentence, title, descriptor) are the most powerful.}
\label{fig:feats}
\end{figure}

\onecolumn

\section{Appendix: Qualitative Visualization}

\begin{figure}[h]
\centering
\includegraphics[width=0.9\textwidth]{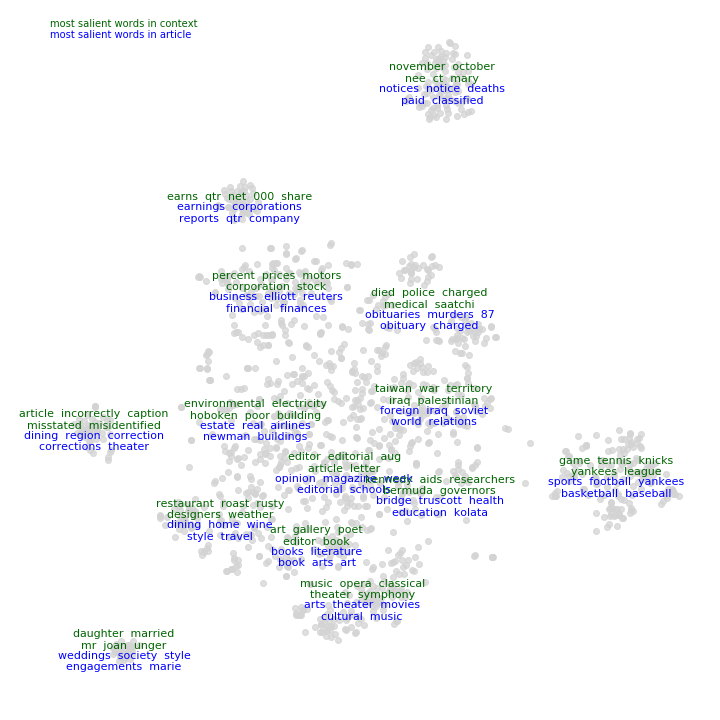} 
\caption{{\em T-SNE plot of context embeddings}. We cluster the {\em first sentence} embedding of 5000 articles and project the 512-d context vectors to two dimensions with t-SNE.
We group context vectors into clusters with k-means and compute TF-IDF scores separately for context (green) and sentence (blue) words and show the top 5 for each.
Notice how the set of five green words cohere with the five blue words, indicating the CUE embeddings project context and metadata to a similar space as the article contents. 
The clustering recovers meaningful news topics, such as company earning reports, obituaries, sports, books and art.}
\label{fig:tsne}
\end{figure}

\end{document}